\documentclass[runningheads]{llncs}

\usepackage{eccv}

\usepackage{eccvabbrv}
\usepackage{cite} 
\usepackage{booktabs}
\usepackage{array}
\usepackage{graphicx}
\usepackage{booktabs}
\usepackage{booktabs}
\usepackage{multirow}   
\usepackage{xcolor}   
\usepackage{amsmath} 
\usepackage[accsupp]{axessibility}  
\usepackage{rotating}   
\usepackage{subcaption}  
\usepackage{booktabs}   
\usepackage{array}     
\usepackage{colortbl}
\usepackage{wrapfig}

\usepackage{hyperref}

\usepackage{orcidlink}

\begin{document}

\title{ScanFocus: A Coarse-to-Fine Framework for Spatio-Temporal Video Grounding} 

\titlerunning{ScanFocus: A Coarse-to-Fine Framework for STVG} 

\author{
Kai Chen\inst{*} \orcidlink{0009-0006-9053-3469} \and
Ming Dai\inst{*} \orcidlink{0009-0004-6133-0035} \and
Wenxuan Cheng \orcidlink{0009-0008-5915-9565} \and
Wankou Yang\inst{\dagger} \orcidlink{0000-0002-6385-6776}
}

\authorrunning{K.~Chen et al.}

\institute{Southeast University
}

\maketitle

\begingroup
\renewcommand{\thefootnote}{\fnsymbol{footnote}}
\footnotetext[1]{Equal contribution.}
\footnotetext[4]{Corresponding author.}
\endgroup

\begin{abstract}
Spatio-Temporal Video Grounding (STVG) aims to retrieve the visual trajectory of a specific object from a video stream as described by a natural language expression. However, most advanced methods struggle to balance global context modeling with precise boundary localization. Due to the prohibitive computational costs of processing long videos, these approaches typically resort to low-rate temporal downsampling and implicit motion modeling. This inevitably suppresses high-frequency boundary cues and neglects the explicit inter-frame dependencies required for precise boundary delineation. To address these limitations, we present \textbf{ScanFocus}, a novel coarse-to-fine framework that decouples the STVG task into a global spatio-temporal scan and a local boundary focus. Specifically, we utilize a unified vision-language fusion encoder combined with a lightweight Deformable Semantic-Motion Fusion module to efficiently align multimodal features and generate coarse proposals. To recover the suppressed fine-grained details, we introduce the Semantic-Guided Temporal Aggregator (SGTA) in the refinement stage. By densely sampling around coarse boundaries, SGTA explicitly models short-term temporal interactions under semantic guidance, capturing rapid motion changes for precise timestamp regression. Extensive experiments on three widely used benchmarks demonstrate the performance superiority of our proposed method over previous  approaches. Code will be released at \url{https://github.com/TenMinutes209/ScanFocus}.
\end{abstract}

\section{Introduction}
\label{sec:intro}

The objective of Spatio-Temporal Video Grounding~\cite{zhang2020where} is to retrieve the visual trajectory of a specific object from a video stream as described by a natural language expression. This dual-domain localization task demands not only robust spatial object detection but also precise temporal boundary delineation, making it a sophisticated frontier for evaluating cross-modal reasoning capabilities in dynamic environments. 

Recently, capitalizing on the powerful attention mechanisms and compact pipelines of Transformers~\cite{attention}, researchers have introduced many Transformer-based architectures  to address this challenge, achieving unprecedented results. By leveraging sophisticated encoder-decoders to comprehensively fuse visual and linguistic features, these methods~\cite{yang2022tubedetr,jin2022embracing, cgstvg,tastvg,su2021stvgbert,stvgformer} have significantly promoted the synergistic recognition of multimodal information and established new state-of-the-art (SOTA) performance in spatio-temporal alignment. 

\begin{figure}[tb]
  \centering
  \includegraphics[height=5.6cm]{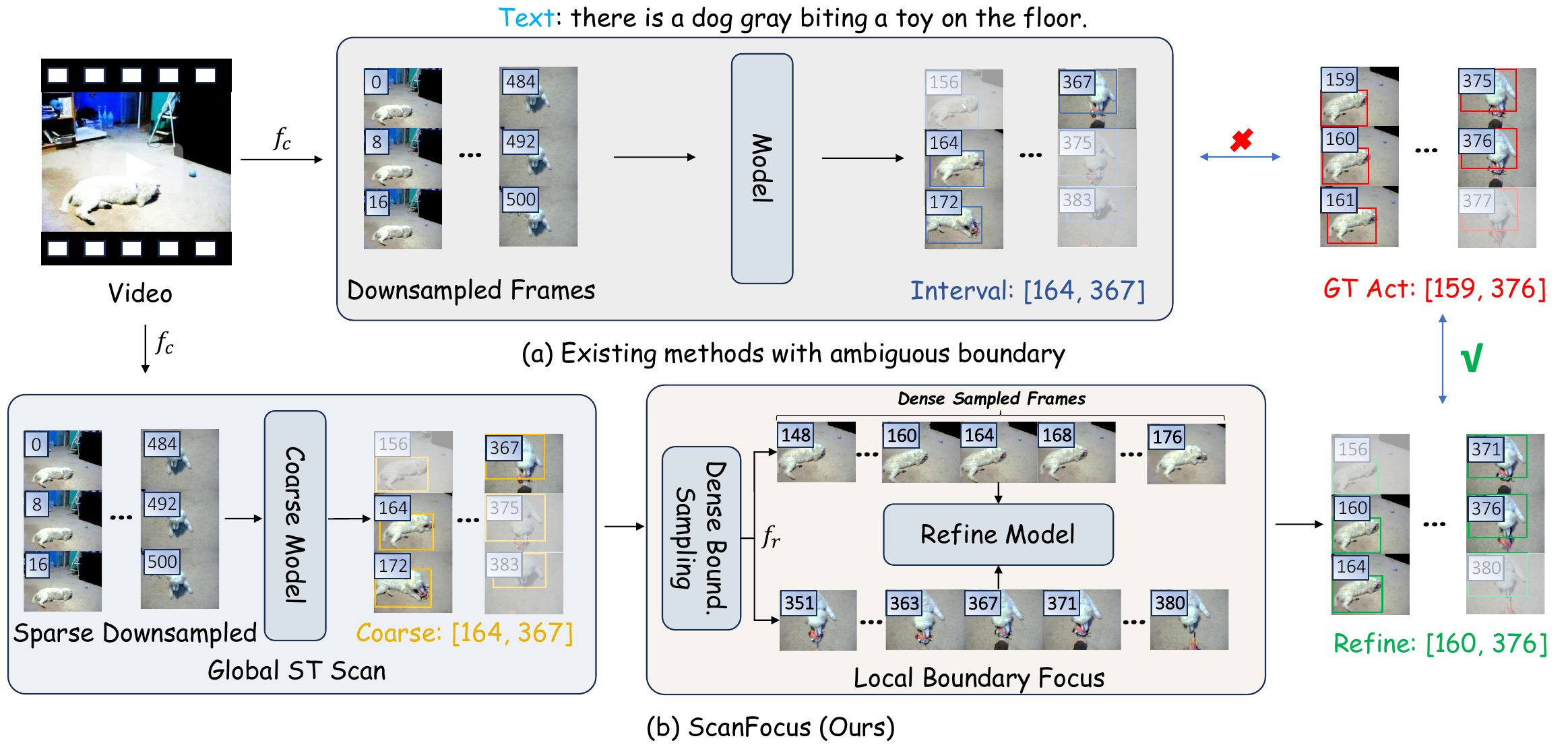}
  \caption{Comparison of temporal boundary localization paradigms.  (a) Existing Transformer-based methods often produce ambiguous boundaries due to the suppression of high-frequency temporal cues caused by global downsampling. (b) Our proposed method adopts a coarse-to-fine framework that first generates a coarse interval at a low frame rate, followed by boundary dense sampling to recover fine-grained details for precise localization.
  }
  \label{fig:motivation}
\end{figure}

Despite these advances, existing Transformer-based STVG methods are inherently limited by the \textbf{suppression of high-frequency cues}. Due to the prohibitive computational costs of processing long video sequences, existing approaches~\cite{yang2022tubedetr, jin2022embracing, cgstvg,tastvg} typically resort to low-rate temporal subsampling designed for global semantic alignment. Consequently, high-frequency temporal boundary cues are discarded, and critical frames corresponding to the ground-truth boundaries are often entirely skipped during sampling. This physically prevents the model from accessing exact boundary features, leading to temporally ambiguous predictions, as visually explicated in Fig.~\ref{fig:motivation}(a).

This ambiguity is further exacerbated by a \textbf{lack of explicit temporal modeling}. To maintain efficiency under global modeling, current methods primarily focus on frame-level cross-modal interaction—fusing appearance, motion, and linguistic features within each individual frame. They rely heavily on motion features implicitly encoded by the backbone, lacking a dedicated mechanism to capture the fine-grained temporal dependencies between frames for precise boundary delineation. Moreover, forcing the encoder to simultaneously accommodate multi-modal features under the static MDETR~\cite{mdetr} paradigm creates a significant optimization bottleneck. This overloaded fusion traps the model in suboptimal solutions, limiting high-precision temporal localization.

\begin{wrapfigure}[19]{r}{0.5\textwidth}
  \centering
  \includegraphics[width=0.43\textwidth]{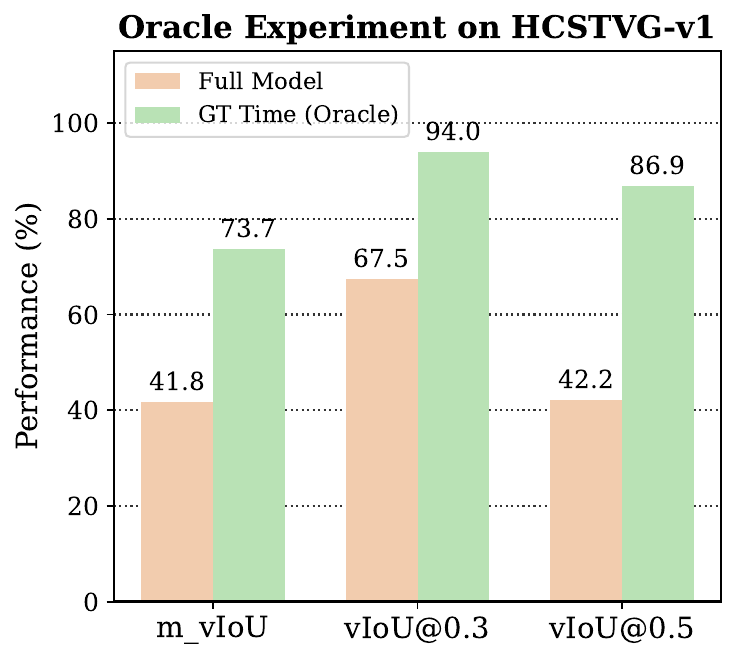}
  \caption{\textbf{Oracle Experiment on HCSTVG-v1.} Using GT timestamps instead of predicted intervals substantially improves performance, identifying temporal localization as the main bottleneck.}
  \label{fig:oracle}
\end{wrapfigure}
To empirically validate this \textbf{temporal localization bottleneck}, we conduct an oracle analysis on the HC-STVGv1 dataset (for other datasets, please kindly refer to Supplementary Material). As illustrated in Fig.~\ref{fig:oracle}, we compare our full model's predictions against an Oracle setting where the spatial tube sequences are evaluated using Ground Truth (GT) timestamps. We observe a staggering performance leap: for instance, the vIoU@0.3 surges from 67.5\% to 94.0\%, and the vIoU@0.5 jumps from 42.2\% to 86.9\%. This substantial gap demonstrates that our framework already possesses highly accurate spatial grounding capabilities, confirming that the primary constraint on STVG is the temporal ambiguity.

To address these limitations, in this paper, we introduce \textbf{ScanFocus}, a novel coarse-to-fine framework that decouples STVG into a global spatio-temporal scan and a local boundary focus. As illustrated in Fig.~\ref{fig:motivation}(b), mirroring the human visual strategy of global scanning followed by local scrutiny, our framework integrates sparse global retrieval with dense boundary refinement. This design effectively recovers suppressed high-frequency cues while establishing explicit inter-frame dependencies for precise temporal delineation. Specifically, we first utilize a unified vision-language fusion encoder combined with a Deformable Semantic-Motion Fusion encoder~\cite{deformable} to efficiently align multimodal features, followed by dual DETR-style decoders~\cite{detr} to generate spatial tubes and coarse temporal intervals, respectively. Furthermore, to capture the high-frequency cues suppressed in the global scan, we introduce a Local Boundary Focus stage. Here, we perform dense boundary sampling around the predicted coarse boundaries and employ a Semantic-Guided Temporal Aggregator module to explicitly model short-term temporal dependencies within local windows. As a result, we recover the  boundary details and refine the coarse proposals into precise start-end timestamps through the refine decoder. To sum up, our contributions include:
\begin{itemize}
    \item We propose \textbf{ScanFocus}, a  coarse-to-fine framework that decouples  STVG into a global spatio-temporal scan and a local boundary focus. This architecture effectively mitigates the suppression of high-frequency cues in global modeling by integrating sparse retrieval with dense boundary refinement.
    
    \item To resolve the lack of explicit inter-frame dependencies in existing methods, we introduce the \textbf{Semantic-Guided Temporal Aggregator} within the refinement stage. This module explicitly models short-term temporal interactions within densely sampled boundary windows, enabling precise regression of start and end timestamps.
    
    \item We conduct extensive experiments on three widely used benchmarks  to demonstrate the performance superiority of our proposed method over previous SOTA approaches, validating the effectiveness of our strategy.
    
\end{itemize}

\section{Related Work}

\subsection{Spatio-Temporal Video Grounding}
Spatio-Temporal Video Grounding~\cite{zhang2020where} aims to precisely localize a target of interest both spatially and temporally within an untrimmed video sequence, based on a given natural language description. Existing STVG methodologies have primarily evolved through several distinct paradigms. Early approaches~\cite{zhang2020where,hcstvg,omrn} typically adopted a two-stage strategy, which first leveraged pre-trained object detectors to generate a redundant set of region proposals and subsequently selected the optimal candidate that best aligns with the textual query. However, these methods are often constrained by the quality of the external detector and suffer from high computational overhead. To mitigate these limitations, subsequent advancements~\cite{su2021stvgbert,yang2022tubedetr,jin2022embracing, stvgformer,vgdino,cgstvg,tastvg} have shifted toward a unified one-stage framework. Inspired by the success of Transformers~\cite{attention}, these methods employ an end-to-end encoder-decoder architecture to directly regress the spatio-temporal tubelets without relying on any pre-defined proposals. By facilitating tighter cross-modal interaction and eliminating the heuristic proposal generation process, this paradigm has established new state-of-the-art performance. Most recently, an emerging paradigm has integrated Multimodal Large Language Models (MLLMs)~\cite{qwen3,gpt4,dsr1,internvl3,seed15} into STVG~\cite{thinking,spacevllm,astvg,llavast}, harnessing their profound reasoning capabilities to further enhance fine-grained spatio-temporal understanding. However, due to the inherent ambiguity of temporal boundaries and the inevitable suppression of high-frequency boundary cues caused by temporal downsampling, existing models frequently struggle with imprecise temporal localization. To address these challenges, we propose a coarse-to-fine strategy to improve the temporal grounding performance.
\subsection{Vision-Language Modeling}
The paradigm of Vision-Language Modeling (VLM) is designed to bridge the semantic gap between the visual and textual modalities, enabling a unified understanding of heterogeneous data. Recently, it has been widely adopted in the field of visual grounding~\cite{seqtr, lvit, mdetr, eevg, segvg, simvg, deris, gc3vg, instancevg}, visual question answering~\cite{vqa,defense, longvlm}, visual captioning~\cite{pixellm, gu2023text, accurate, imagecaption}, video temporal grounding~\cite{vtg,moment-detr, qd-detr, cg-detr,univtg, r2tuning, flashvtg, tall, multilevel, zhang2020learning, momentseg}, etc. In STVG, current mainstream frameworks~\cite{yang2022tubedetr, cgstvg, tastvg} heavily rely on the MDETR~\cite{mdetr} pipeline for cross-modal fusion. Despite their success, these methods typically struggle with an optimization bottleneck caused by forcing a spatially-aligned encoder to simultaneously model both temporal dynamics and motion cues. Such a paradigm fails to explicitly decouple the disparate feature spaces required for spatial and temporal grounding, leading to sub-optimal performance due to the compromised representation capability of the overloaded multimodal encoder. In contrast, our approach eliminates the reliance on this complex encoder-decoder architecture. By leveraging a vision-language fusion encoder~\cite{beit, beit3} to establish robust vision-language alignments, we reformulate the intricate tri-modal interaction into a streamlined cross-modal fusion task.

\subsection{Video Temporal Grounding}
The objective of video temporal grounding (VTG)~\cite{vtg} is to establish a fine-grained alignment between linguistic descriptions and their corresponding temporal segments in untrimmed videos. VTG has been extensively studied through its two sub-tasks: moment retrieval (MR) and highlight detection (HD). Early methodologies~\cite{tall,multilevel,zhang2020learning} primarily relied on proposal-based paradigms, which first generated candidate segments via sliding windows or anchor mechanisms and then performed cross-modal ranking. While effective, these approaches often suffer from high redundancy and struggle with coarse temporal boundaries due to their reliance on predefined heuristics. In contrast, recent advancements have predominantly shifted toward end-to-end DETR-based frameworks~\cite{moment-detr, qd-detr, cg-detr,locvtp, endvtg, lgvtg, canshuffle, protege, tpvtg, univtg, barrios2023localizing}. By utilizing learnable queries to directly regress temporal boundaries, these methods eliminate the need for hand-crafted proposals and facilitate global cross-modal reasoning. Subsequent studies, such as R2-Tuning~\cite{r2tuning} and FlashVTG~\cite{flashvtg}, further refined this paradigm by incorporating multi-scale temporal modeling to capture events of diverse durations. However, STVG~\cite{zhang2020where} presents a more formidable challenge than VTG, as it extends the grounding task from a single temporal axis to a joint spatio-temporal manifold, requiring a synergistic understanding of both object appearance and motion evolution.

\section{Method}
In this section, we first formulate the spatio-temporal video grounding task and review the prevalent MDETR-based framework in Sec.~\ref{sec:preliminary}. In Sec.~\ref{sec:overview}, we  give an overview of our framework. Then, we  detail the Global  Spatio-Temporal Scan in Sec.~\ref{sec:coarse} and the Local Boundary Focus in Sec.~\ref{sec:refine}. In Sec.~\ref{sec:opt}, we  describe the training objectives and optimization strategy.

\begin{figure}[tb]
  \centering
  \includegraphics[height=7cm]{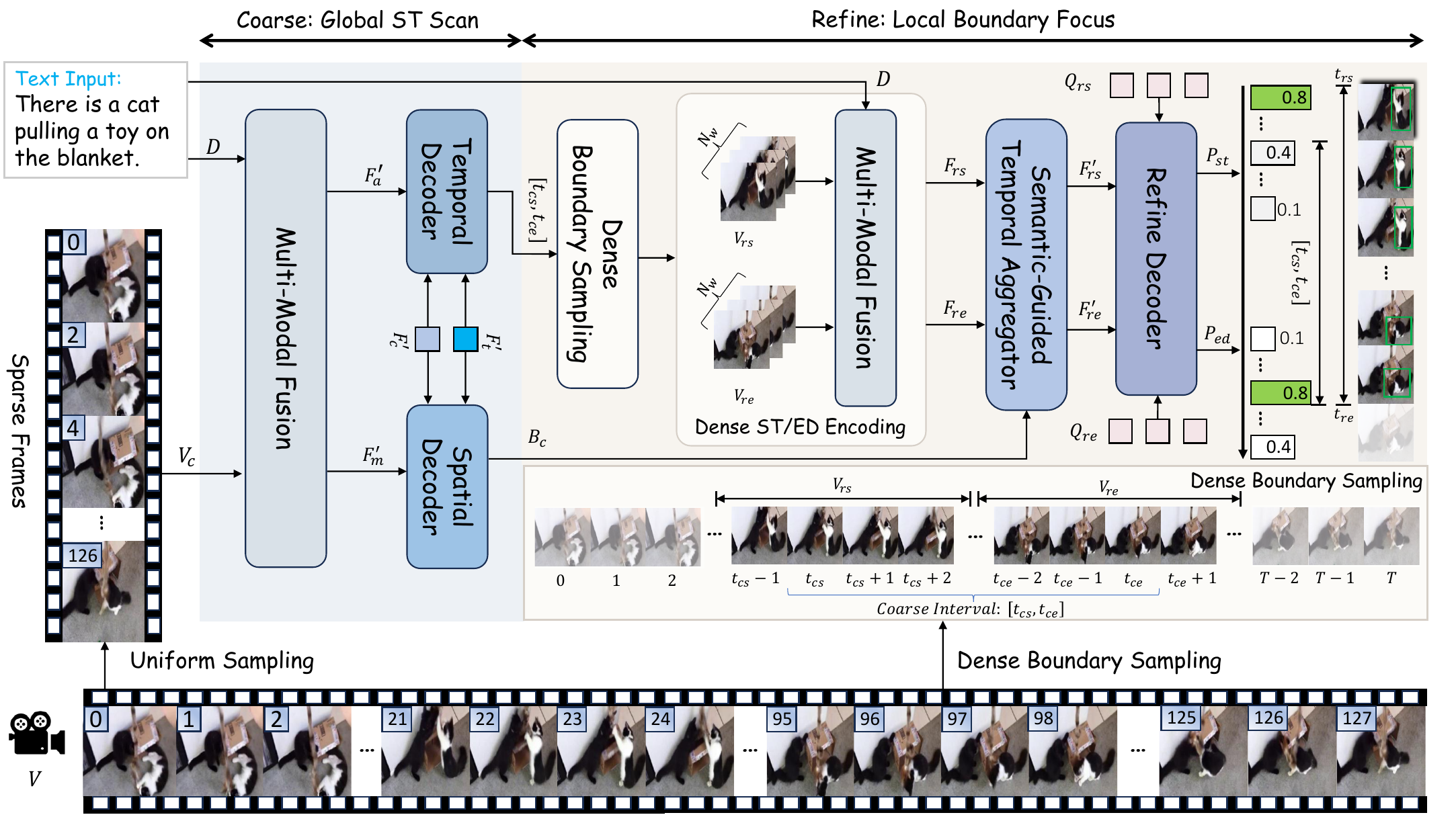}
  \caption{
  \textbf{Overview architecture of our proposed ScanFocus.}
  The framework follows a coarse-to-fine paradigm, decoupling the task into two stages:
  \textbf{1) Global  Spatio-Temporal Scan:} We first utilize a unified vision-language fusion encoder combined with a lightweight Semantic-Motion Fusion Encoder to efficiently align multimodal features. Dual DETR-style decoders are then employed to generate coarse spatial tubes and temporal intervals.
  \textbf{2) Local Boundary Focus:} To recover high-frequency cues suppressed in the coarse stage, we perform dense sampling around the predicted coarse boundaries. The Semantic-Guided Temporal Aggregator explicitly models short-term dependencies within these local windows, which are finally fed into dual refine decoders to predict precise start and end timestamps.
}
  \label{fig:framework}
\end{figure}

\subsection{Preliminary}
\label{sec:preliminary}
\textbf{Formulation. } Given a video sequence $V \in \mathbb{R}^{T \times H \times W \times 3}$ and a natural language query $D$, the goal of STVG is to precisely output a spatio-temporal tube (a bounding box sequence with temporal boundaries) grounding $D$ in $V$. 

\begin{figure}[tb]
  \centering
  \includegraphics[width=\linewidth]{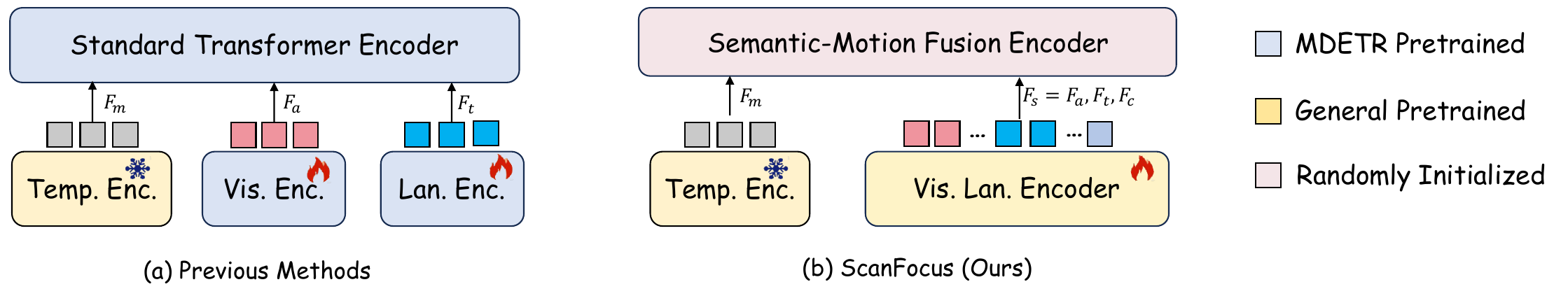} 
  \caption{\textbf{Comparison of multi-modal fusion mechanisms.} 
  \textbf{(a)} Existing methods fuse vision, language, and motion simultaneously. This fully-coupled tri-modal alignment is inherently challenging and yields suboptimal representations on limited downstream datasets. 
  \textbf{(b)} We adopt a decoupled strategy leveraging a general vision-language encoder to extract visual and linguistic features pre-aligned on massive upstream datasets, followed by a lightweight Semantic-Motion Fusion Encoder to further align these robust semantic representations with motion dynamics.}
  \label{fig:fusion_comparison}
\end{figure}

As illustrated in Fig.~\ref{fig:fusion_comparison}(a), prevailing STVG frameworks~\cite{yang2022tubedetr, jin2022embracing,cgstvg, tastvg} typically build upon the MDETR~\cite{mdetr} architecture. They first extract static appearance ($F_a$), linguistic ($F_t$), and temporal motion ($F_m$) features using independent MDETR pre-trained backbones. Subsequently, a heavy MDETR pre-trained standard transformer encoder is applied to project and couple these heterogeneous representations within a shared latent space:
\begin{equation}
\hat{F}_a, \hat{F}_m, \hat{F}_t = \text{Encoder}(\text{Concat}(\text{Proj}(F_a, F_m, F_t))),
\end{equation}
where $\text{Proj}(\cdot)$ denotes linear projections for channel alignment. 

Finally, task-specific decoders are employed to predict the spatio-temporal locations. The MDETR pre-trained spatial locator regresses bounding box sequences $\mathbf{B}$, while an auxiliary temporal decoder concurrently predicts the temporal boundaries $(s, e)$.

\subsection{Overview}
\label{sec:overview}
To address the limitations described above, we propose \textbf{ScanFocus}. Fig.~\ref{fig:framework} illustrates the overall architecture. Different from previous methods that rely on heavy static pre-training, ScanFocus adopts a coarse-to-fine philosophy. The pipeline begins with a unified vision-language fusion encoder initialized by video-language pre-training to extract aligned features, followed by a lightweight Deformable Semantic-Motion Encoder to efficiently fuse semantic and motion information. 
The encoded features are then processed in two cascaded stages: 
1) The \textbf{Global  Spatio-Temporal Scan} stage (Sec.~\ref{sec:coarse}), which utilizes dual decoders to generate coarse spatial tubes and temporal intervals from sparse global queries; 
2) The \textbf{Local Boundary Focus} stage (Sec.~\ref{sec:refine}), which performs dense boundary sampling and employs a Semantic-Guided Temporal Aggregator to capture fine-grained inter-frame dependencies for precise boundary refinement. 

\subsection{Coarse: Global  Spatio-Temporal Scan}
\label{sec:coarse}
\textbf{Feature Extraction.} As shown in Fig.~\ref{fig:framework}, given the redundancy in adjacent video frames and the need for efficient global modeling, we first perform a \textit{sparse sampling} strategy on the input video. Specifically, we uniformly sample $T_c$ frames from the original video sequence with a low sampling rate $f_c$ to construct the coarse input, denoted as $V_c \in \mathbb{R}^{T_c \times H \times W \times 3}$.

Subsequently, to bypass the optimization bottleneck of task-specific pre-training (Fig.~\ref{fig:fusion_comparison}(a)), we leverage general pre-trained foundation models for feature extraction, as illustrated in Fig.~\ref{fig:fusion_comparison}(b), we adopt a unified Vision-Language Fusion encoder (VLF) to extract appearance and linguistic features from these sampled frames. Unlike traditional vision-language paradigms that employ separate unimodal encoders to extract independent appearance and linguistic features followed by a cross-modal fusion module, our VLF is inherently unified. Pre-trained on massive-scale image-text pairs, it implicitly aligns visual and textual representations within a shared semantic space during the encoding phase, eliminating the domain discrepancy often observed in late-fusion architectures. Specifically, we use BEiT-3~\cite{beit3} to extract appearance, linguistic, and [CLS] features, denoted as ${F}_a \in \mathbb{R}^{T_c \times N_a \times C_a}$, ${F}_l \in \mathbb{R}^{T_c \times L \times C_a}$, and ${F}_c \in \mathbb{R}^{T_c \times 1 \times C_a}$, where $N_a$ and $L$ denote the token counts, and $C_a$ represents the feature dimension.

For motion feature extraction, following previous methods in~\cite{cgstvg, tastvg}, we introduce an video encoder to explicitly extract motion features. Specifically, we employ a frozen pre-trained VideoMAE~\cite{videomae} to generate the motion representation ${F}_m \in \mathbb{R}^{T \times N_m \times C_m}$, where $N_m$ and $C_m$ are the motion token count and channel dimension, respectively.
\subsubsection{Semantic-Motion Fusion.} Previous STVG methods~\cite{cgstvg, tastvg, yang2022tubedetr, jin2022embracing} often rely on heavy, fully-coupled Transformer encoders for multi-modal interaction. In contrast, leveraging the inherent cross-modal alignment of our VLF encoder, we propose a lightweight Deformable Semantic-Motion Fusion module that reformulates this interaction as a multi-scale deformable attention task. Specifically, we project all features into a shared dimension $C$ to form a multi-level feature pyramid $\mathcal{X} = \{F_a, F_m, F_t, F_c\}$, and define the concatenated query sequence as $\mathbf{Z} \in \mathbb{R}^{N_{total} \times C}$. For each query token $z_q \in \mathbf{Z}$, the fused representation is computed by sparsely sampling from $\mathcal{X}$:
\begin{equation}
\text{Fusion}(z_q) = \sum_{l=1}^{4} \sum_{k=1}^{K} A_{lqk} \cdot \mathbf{W}_v \cdot \Phi(\mathcal{X}^l; p_q + \Delta p_{lqk}), 
\end{equation}
where $p_q$, $\Delta p_{lqk}$, and $A_{lqk}$ denote the reference point, learnable offset, and attention weight, respectively. This mechanism enables linguistic tokens to sparsely attend to salient motion or appearance regions with linear complexity. Finally, after $L_{enc}$ layers of stacked fusion, the unified sequence is disentangled into enriched modality streams:
\begin{equation}
  {F_{a}^{'}, F_{m}^{'}, F_{t}^{'}, F_{c}^{'}} = \text{Split}\left(\Phi_{\text{D}}\left(\text{Concat}(F_a, F_m, F_t, F_c)\right)\right),
\end{equation}
where $\Phi_{\text{D}}$ denotes the stacked deformable encoder layers. $\text{Split}(\cdot)$ recovers the individual modality streams based on their original token lengths.

\subsubsection{Global Spatio-Temporal Decoder.} Following established practices~\cite{cgstvg, tastvg}, we employ separate spatial and temporal decoders. To accelerate convergence and improve localization, we propose a semantic-injection strategy to explicitly incorporate linguistic and global contexts into the queries prior to decoding. Specifically, we initialize learnable queries $Q_s, Q_t \in \mathbb{R}^{T_c \times C}$ for the spatial and temporal branches. We then sequentially inject semantic information from the text features $F_{t}^{'}$ and the global context token $F_{c}^{'}$ into these queries via standard cross-attention layers, yielding semantically conditioned queries $Q^{i}_s$ and $Q^{i}_t$. This initialization ensures that the queries are target-aware before interacting with the visual ($F_{a}^{'}$) and motion ($F_{m}^{'}$) features. With these conditioned queries, the decoders perform iterative reasoning through $K$ stacked layers. Finally, task-specific prediction heads generate the coarse outputs: a 3-layer MLP predicts the bounding box sequence $\mathbf{B}_{c} \in \mathbb{R}^{T_c \times 4}$, while a temporal head outputs frame-wise start/end probabilities to form the coarse temporal interval $\tau_{c} = (t_{cs}, t_{ce})$.

\subsection{Refine: Local Boundary Focus}\label{sec:refine}\textbf{Dense Boundary Sampling.} To recover the fine-grained motion details that suppressed by the sparse global scan, we introduce a boundary-focused dense sampling strategy. Given the predicted coarse interval $\tau_{c}$, we construct two local observation windows centered at $t_{cs}$ and $t_{ce}$. Distinct from the global stage, we sample $N_w$ frames within each window using a high sampling rate $f_r = k_r \cdot f_c$ (where $k_r > 1$ is the density factor) to ensure dense temporal coverage. This yields two dense frame sequences: $\mathcal{V}_{rs}$ and $\mathcal{V}_{re}$, both of shape $\mathbb{R}^{N_w \times H \times W \times 3}$. 

To maintain semantic consistency and maximize parameter efficiency, we strategically reuse the feature extraction and fusion modules from the coarse stage (Sec.~\ref{sec:coarse}). Specifically, $\mathcal{V}_{rs}$ and $\mathcal{V}_{re}$ are independently processed by the shared vision-language fusion encoder, Video Encoder, and the  Semantic-Motion Fusion module. This yields refined multi-modal feature sequences ${F}_{rs}, {F}_{re}$, encoding rich boundary details for precise delineation.

\begin{wrapfigure}{r}{0.48\textwidth}
  \centering
  \includegraphics[width=\linewidth]{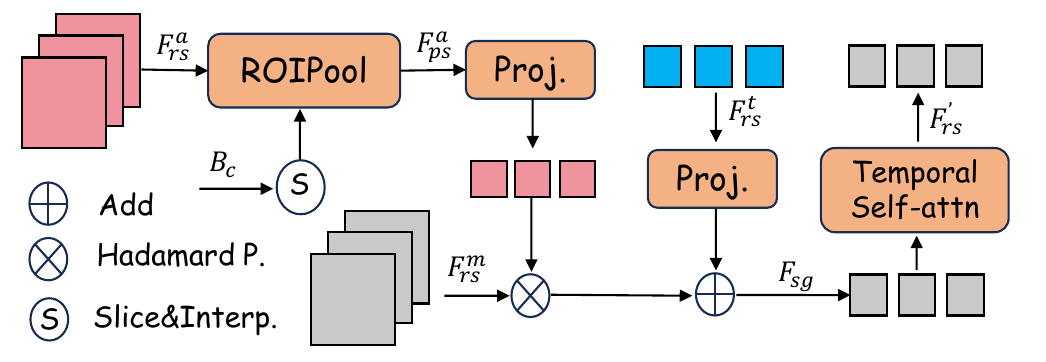}
  \caption{\textbf{Detailed architecture of SGTA module.}}
  \label{fig:sgta}
\end{wrapfigure}

\subsubsection{Semantic-Guided Temporal Aggregator.} While the shared encoder effectively extracts frame-wise representations, precise boundary delineation inherently requires capturing fine-grained temporal evolution, which relies on explicit inter-frame reasoning. To bridge this gap, we introduce SGTA. Leveraging the reduced temporal footprint of the local windows ($N_w$), SGTA performs efficient, dense temporal modeling to capture high-frequency motion cues omitted in the coarse stage. As shown in Fig. \ref{fig:sgta}, taking the start branch as an example, SGTA first generates a dense spatial prior for the local window. Given the coarse bounding box sequence $\mathbf{B}_{c}$ and the predicted start timestamp $t_{cs}$, we retrieve the subset of coarse boxes corresponding to the temporal span of the refinement window. Since this window contains $N_w$ frames sampled at a high rate $f_r$, we perform temporal linear interpolation on this sparse subset to upsample it to the target length $N_w$. This efficiently generates the dense spatial prior $\mathbf{B}_{st} \in \mathbb{R}^{N_w \times 4}$ without requiring heavy external object detectors. With the refined feature sequence $F_{rs}$ obtained from the shared encoder, we isolate its distinct modality streams: appearance ($F_{rs}^a$), motion ($F_{rs}^m$), and text ($F_{rs}^t$). To rigorously align the visual representations with the tracked object, we employ RoI Pooling on the appearance features guided by the spatial prior $\mathbf{B}_{st}$. This yields the object-aligned appearance sequence $F_{ps}^a$, which focuses exclusively on the target regions:
\begin{equation}
  F_{ps}^a = \text{ROIPool}(F_{rs}^a, \mathbf{B}_{st}).
\end{equation}

Subsequently, to explicitly align the temporal dynamics with the target object and linguistic query, we utilize these semantic features to guide the motion representation. To filter out irrelevant background dynamics, the motion features $F_{rs}^m$ are modulated by the object-aligned appearance features $F_{ps}^a$, effectively highlighting motion patterns associated with the visual referent. Furthermore, the text features $F_{rs}^t$ are injected as a linguistic bias. The final semantic-guided motion feature $F_{sg}$ is formulated as:
\begin{equation}
  F_{sg} = \left( \phi_{a}(F_{ps}^a) \odot F_{rs}^m \right) \oplus \phi_{t}(F_{rs}^t),
\end{equation}
where $\odot$ and $\oplus$ denote the Hadamard product and element-wise addition, respectively. $\phi_{a}$ and $\phi_{t}$ represent learnable linear projections. Finally, to capture high-frequency cues and model explicit inter-frame dependencies, we flatten $F_{sg}$ into a unified sequence of length $N_w \times N_m$ (where $N_m$ denotes the number of motion tokens) and process it through a stack of $L_t$ temporal self-attention layers:
\begin{equation}
F_{rs}^{'} = \text{MHSA}_{\times L_t}(\text{Flatten}(F_{sg})).
\end{equation}

This flattening structure ensures that all tokens within the local window are mutually perceptible, enabling fine-grained global reasoning across the boundary region. The resulting $F_{rs}^{'}$ effectively aggregates dense boundary details, serving as the input for the subsequent prediction heads.

\subsubsection{Refine Decoder.} Similar to the coarse stage, a DETR-style decoder $\text{Dec}_{rs}$ is employed to probe the aggregated features $F_{rs}^{'}$. The decoded embeddings are then directly projected by task-specific prediction heads to generate the frame-wise boundary probability $\mathbf{P}_{st}$ and the auxiliary action confidence $\mathbf{A}_{st}$. The end branch follows an identical procedure to obtain $\mathbf{P}_{ed}$ and $\mathbf{A}_{ed}$.

\subsection{Optimization}\label{sec:opt}To ensure optimization stability and prevent fine-grained gradients from perturbing well-learned global priors, we adopt a decoupled two-stage training paradigm. 
For coarse stage, we first train the global modules (backbone, fusion encoder, and coarse decoders) using $\mathcal{L}_{c}$. The spatial decoder is supervised by L1 and GIoU losses given the ground-truth box sequence $\mathbf{B}^*$, while the temporal decoder uses KL divergence to align the predicted distributions $\hat{P}_{s/e}$ with Gaussian-smoothed boundary labels $t_{s/e}^*$:
\begin{equation} \mathcal{L}_{c} = \lambda_{box} (\mathcal{L}_{L1}(\mathbf{B}_{c}, \mathbf{B}^*) + \mathcal{L}_{IoU}(\mathbf{B}_{c}, \mathbf{B}^*)) + \lambda_{tmp} (\mathcal{L}_{KL}(\hat{P}_{s}, t_s^*) + \mathcal{L}_{KL}(\hat{P}_{e}, t_e^*)).
\end{equation}

For refine stage, we freeze the coarse stage parameters and exclusively optimize the refinement modules (SGTA and Refine Decoders). We formulate boundary localization and action detection as binary classification tasks, optimized via Binary Cross-Entropy (BCE) loss:
\begin{equation} \mathcal{L}_{r} = \lambda_{ref} \mathcal{L}_{BCE}(\mathbf{P}_{st/ed}, \mathbf{y}_{st/ed}^*) + \lambda_{act} \mathcal{L}_{BCE}(\mathbf{A}_{st/ed}, \mathbf{y}_{act}^*),
\end{equation}
where $\mathbf{P}_{st/ed}$ and $\mathbf{A}_{st/ed}$ are the predicted boundary and action probabilities, and $\mathbf{y}^*$ are the corresponding binary ground-truth labels. The hyperparameters $\lambda$ balance their respective loss terms.
\section{Experiments}
\subsection{Experiment Settings}
\textbf{Datasets.} The commonly used datasets in spatio-temporal video grounding are HC-STVGv1/v2 and VidSTG. HC-STVGv1 comprises 5,660 untrimmed video clips, which are partitioned into 4,500 training and 1,160 testing video-text pairs. HC-STVGv2 significantly expands upon this scale to include a total of 16,544 samples, subdivided into 10,131 samples for training, 2,000 for validation, and 4,413 for testing. Due to the unavailability of ground-truth annotations for the test set, we report results on the validation set, following prior work. VidSTG provides a more substantial corpus for evaluation, encompassing 99,943 video-text pairs derived from 5,436 source videos. This dataset is characterized by its high linguistic diversity, featuring a mix of 44,808 declarative sentences and 55,135 interrogative queries. The data is organized into training, validation, and testing subsets with 80,684, 8,956, and 10,303 sentences respectively, which are mapped across 5,436, 602, and 732 unique video sequences.
\subsubsection{Metrics.} Following prior work, we employ m\_tIoU, m\_vIoU, and vIoU@R as evaluation metrics. m\_tIoU measures the ability of temporal grounding by computing the average temporal Intersection-over-Union (tIoU) score across all the test set. m\_vIoU assesses spatial grounding quality by computing the average IoU across space and time between predicted and annotated spatio-temporal tubes, and vIoU@R measures the performance using ratios of samples with vIoU greater than $R$ in test sets. For detailed metrics, please kindly refer to \cite{yang2022tubedetr}.

\subsubsection{Implementation.} Our model employs BEiT-3~\cite{beit3} for vision-language modeling and VideoMAE~\cite{videomae} for motion representation. Inputs are resized to $384 \times 384$. We sample 64 coarse frames and 8 refine frames ($k_r\!=\!2$). Detailed network configurations and loss weights are provided in the Supplementary Material.

\subsection{Quantitative Results}
\textbf{Results on HC-STVGv1/v2.} 
To validate the efficacy of our method, we compare it against state-of-the-art approaches on the HC-STVGv1 and v2 datasets. Table~\ref{tab:sota_comparison_v1} and Table~\ref{tab:sota_comparison_v2} present the comparison results on the respective test sets.

As shown in Table~\ref{tab:sota_comparison_v1}, ScanFocus significantly outperforms existing methods across all metrics on HC-STVGv1. Notably, on the strict localization metrics vIoU@0.3 and vIoU@0.5, our method achieves 67.5\% and 42.2\%, surpassing the previous best method TA-STVG~\cite{tastvg} by \textbf{4.4\%} and \textbf{5.4\%}, respectively. Similarly, on the HC-STVGv2 dataset (Table~\ref{tab:sota_comparison_v2}), ScanFocus establishes new state-of-the-art performance, achieving a 2.0\% improvement in tiou and a 2.6\% improvement in vIoU@0.5 over TA-STVG. These consistent improvements demonstrate the superiority of our coarse-to-fine paradigm in recovering high-frequency boundary cues for precise temporal delineation.

\begin{table}[t]
  \centering
  \small 
  
  \begin{minipage}{0.48\textwidth}
    \centering
    \caption{Comparison on HC-STVGv1. The best results are highlighted in \textbf{bold}.}
    \label{tab:sota_comparison_v1}
    \resizebox{\textwidth}{!}{ 
      \begin{tabular}{lcccc}
        \toprule
        Methods & m\_tIoU & m\_vIoU & vIoU@0.3 & vIoU@0.5 \\
        \midrule
        STVGBert~\cite{su2021stvgbert}   & -       & 20.4 & 29.4 & 11.3 \\
        TubeDETR~\cite{yang2022tubedetr} & 43.7   & 32.4 & 49.8 & 23.5 \\
        STCAT~\cite{jin2022embracing}     & 49.4   & 35.1 & 57.7 & 30.1 \\
        SGFDN~\cite{wang2023efficient}   & 46.9   & 35.8 & 56.3 & 37.1 \\
        STVGFormer~\cite{stvgformer}  & -     & 36.9 & 62.2 & 34.8 \\
        VG-DINO~\cite{vgdino} & - & 38.3 & 62.5 & 36.1\\
        CG-STVG~\cite{cgstvg}   & 52.8   & 38.4 & 61.5 & 36.3 \\
        TA-STVG~\cite{tastvg} & 53.0 & 39.1 & 63.1 & 36.8\\
        \hline
        \rowcolor{gray!15} \textbf{ScanFocus (Ours)} & \textbf{55.5} & \textbf{41.8} & \textbf{67.5} & \textbf{42.2}\\
        \bottomrule
      \end{tabular}
    }
  \end{minipage}
  \hfill 
  \begin{minipage}{0.48\textwidth}
    \centering
    \caption{Comparison on HC-STVGv2. The best results are highlighted in \textbf{bold}.}
    \label{tab:sota_comparison_v2}
    \resizebox{\textwidth}{!}{
      \begin{tabular}{lcccc}
        \toprule
        Methods & m\_tIoU & m\_vIoU & vIoU@0.3 & vIoU@0.5 \\
        \midrule
        PCC~\cite{pcc}   & -       & 30.0 & - & - \\
        2D-Tan~\cite{2dtan}   & -       & 30.4 & 50.4 & 18.8 \\
        MMN~\cite{mmn} & -       & 30.3 & 49.0 & 25.6 \\
        TubeDETR~\cite{yang2022tubedetr} & 53.9   & 36.4 & 58.8 & 30.6 \\
        STVGFormer~\cite{stvgformer}  & 58.1     & 38.7 & 65.5 & 33.8 \\
        VG-DINO~\cite{vgdino} & - & 39.9 & 67.1 & 34.5\\
        CG-STVG~\cite{cgstvg}   & 60.0   & 39.5 & 64.5 & 36.3 \\
        TA-STVG~\cite{tastvg} & 60.4 & 40.2 & 65.8 & 36.7\\
        \hline
        \rowcolor{gray!15} \textbf{ScanFocus (Ours)} & \textbf{62.4} & \textbf{41.7} & \textbf{68.4} & \textbf{39.3}\\
        \bottomrule
      \end{tabular}
    }
  \end{minipage}
\end{table}
\subsubsection{Results on VidSTG.} To further evaluate the generalization capability of ScanFocus, we conduct experiments on the VidSTG dataset, which is significantly more challenging due to its diverse linguistic expressions (i.e., declarative and interrogative sentences) and complex temporal variations. As summarized in Table~\ref{tab:vidstg_sota}, ScanFocus consistently outperforms existing state-of-the-art methods across both sentence types.
\begin{table}[tb]
  \centering
  \caption{Comparison on VidSTG. The best results are highlighted in \textbf{bold}.}
  \label{tab:vidstg_sota}
  \resizebox{\textwidth}{!}{
  \begin{tabular}{lcccc|cccc} 
    \toprule
    \multirow{2}{*}{Methods} & \multicolumn{4}{c|}{Declarative Sentences} & \multicolumn{4}{c}{Interrogative Sentences} \\ 
    \cmidrule(lr){2-5} \cmidrule(lr){6-9}
    &  m\_tIoU & m\_vIoU & vIoU@0.3 & vIoU@0.5 & m\_tIoU & m\_vIoU & vIoU@0.3 & vIoU@0.5 \\
    \midrule
    STGRN~\cite{zhang2020where} & 48.5 & 19.8 & 25.8 & 14.6 & 47.0 & 18.3 & 21.1 & 12.8 \\
    OMRN~\cite{omrn} & 50.7 & 23.1 & 32.6 & 16.4 & 49.2 & 20.6 & 28.4 & 14.1 \\
    STGVT~\cite{hcstvg} & - & 21.6 & 29.8 & 18.9 & - & - & - & - \\
    STVGBert~\cite{su2021stvgbert} & - & 24.0 & 30.9 & 18.4 & - & 22.5 & 26.0 & 16.0 \\
    TubeDETR~\cite{yang2022tubedetr} & 48.1 & 30.4 & 42.5 & 28.2 & 46.9 & 25.7 & 35.7 & 23.2 \\
    SGFDN~\cite{wang2023efficient} & 45.1 & 28.3 & 41.7 & 29.1 & 44.8 & 25.8 & 36.9 & 23.9 \\
    STVGFormer~\cite{stvgformer} & - & 33.7 & 47.2 & 32.8 & - & 28.5 & 39.9 & 26.2 \\
    STCAT~\cite{jin2022embracing} & 50.8 & 33.1 & 46.2 & 32.6 & 49.7 & 28.2 & 39.2 & 26.6 \\
    CG-STVG~\cite{cgstvg} & 51.4 & 34.0 & 47.7 & 33.1 & 49.9 & 29.0 & 40.5 & 27.5 \\
    TA-STVG~\cite{tastvg} & 51.7 & 34.4 & 48.2 & 33.5 & 50.2 & 29.5 & 41.5 & 28.0 \\
    SpaceVLLM-7B~\cite{spacevllm} & 47.7 & 27.4 & 39.1 & 26.2 & 48.5 & 25.4 & 35.9 & 22.2 \\
    ASTG~\cite{astvg} & 45.6 & 29.2 & 40.3 & 27.8 & -- & -- & -- & -- \\
    \hline
    \rowcolor{gray!15} \textbf{ScanFocus (Ours)} & \textbf{53.3} & \textbf{36.5} & \textbf{50.6} & \textbf{36.0} & \textbf{51.4} & \textbf{30.7} & \textbf{42.5} & \textbf{29.4} \\
    \bottomrule
  \end{tabular}
  }
\end{table}
\subsection{Ablation Studies}
In this section, we perform ablation experiments to investigate the impact of key components in ScanFocus. Unless otherwise specified, all experiments are conducted on the HC-STVGv1 dataset with an input resolution of $224 \times 224$. For brevity, tiou and viou are used to represent m\_tIoU and m\_vIoU.

\subsubsection{Component-wise Effectiveness. }As shown in Table~\ref{tab:ablation_component}, the baseline global scan model achieves a tIoU of 50.9\%. Simply introducing the Dense Sampling (DS) strategy with refine decoders improves tIoU to 52.4\% by recovering high-frequency temporal cues; however, the slight drop in vIoU@0.5 suggests that raw dense frames may introduce redundancy without explicit modeling. Finally, incorporating the proposed SGTA module yields the best performance across all metrics, demonstrating its ability to effectively aggregate dense boundary information through semantic-guided reasoning.

\begin{table}[t]
  \centering
  \small 
 
  \begin{minipage}{0.55\textwidth}
    \centering
    \caption{\textbf{Component-wise effectiveness analysis.} Evaluated on HC-STVGv1. DS means refine stage with multi-modal fusion and start/end decoders.}
    \label{tab:ablation_component}
    \setlength{\tabcolsep}{3pt} 
    \begin{tabular}{lcccc}
      \toprule
      Method & tIoU & vIoU & vIoU@0.3 & vIoU@0.5 \\
      \midrule
      Coarse & 50.9 & 38.2 & 60.7 & 38.0 \\
      + DS & 52.4 & 39.1 & 62.6 & 37.9\\
      + SGTA & \textbf{53.7} & \textbf{40.0} & \textbf{64.0} & \textbf{39.5} \\
      \bottomrule 
    \end{tabular}
  \end{minipage}
  \hfill
  \begin{minipage}{0.40\textwidth}
    \centering
    \caption{\textbf{Impact of Window Size $N_w$.} Performance of ScanFocus varies with different window sizes.}
    \label{tab:windowsize}
    \setlength{\tabcolsep}{5pt} 
    \begin{tabular}{lccc}
      \toprule
      $N_w$ & tIoU & vIoU & GFLOPs \\
      \midrule
      4  & 52.7 & 39.2 & 10.7 \\
      6  & 53.2 & 39.6 & 16.1 \\
      \textbf{8}  & \textbf{53.7} & \textbf{40.0} & \textbf{21.5} \\
      10 & 53.6 & 39.9 & 26.9 \\
      12 & 53.5 & 39.8 & 32.2 \\
      \bottomrule
    \end{tabular}
  \end{minipage}
\end{table}
\subsubsection{Impact of Refine Window Size $N_w$. }
We evaluate the influence of refine window size $N_w \in \{4, 6, 8, 10, 12\}$ to find the optimal temporal receptive field for boundary regression. Given the high sampling rate $f_r$, this window size $N_w$ determines the balance between local detail and temporal span. 
As shown in Table~\ref{tab:windowsize}, the tIoU and vIoU initially increases and peaks at $N_w=8$. This trend suggests that a moderate window is essential for capturing sufficient temporal context to disambiguate the transition between action and background. However, performance begins to degrade when $N_w$ exceeds 8. This degradation is likely because overly long windows introduce redundant background frames and irrelevant motion noise, which dilutes the boundary-specific semantic features within the SGTA module. In terms of efficiency, GFLOPs increase linearly with $N_w$, confirming that our decoupled design maintains stable computational scalability. Given that $N_w=8$ achieves the best trade-off between localization precision and resource consumption, we adopt it as our default configuration.

\subsubsection{Components of SGTA Design. }
To further investigate the internal mechanism of the Semantic-Guided Temporal Aggregator, we perform an exhaustive ablation study on its key components: the Semantic Guide and the Temporal Attention mechanism. The results are summarized in Table~\ref{tab:ablation_sgta}.
As observed, removing the Temporal Attention component leads to a noticeable performance degradation, with tIoU dropping from 53.7\% to 52.8\%. This decline underscores the necessity of explicit inter-frame dependency modeling for capturing long-range temporal dynamics in video sequences. Furthermore, the absence of the Semantic Guide results in a decrease across all metrics, confirming that semantic cues play a pivotal role in filtering out background noise and focusing the model on task-relevant temporal boundaries. The full SGTA configuration achieves the best performance, validating that the synergy between semantic grounding and temporal relation modeling is essential for precise boundary localization.


\subsubsection{Effectiveness of Deformable Semantic-Motion Fusion. }
As shown in Table~\ref{table:fusion_comparison}, our Deformable Semantic-Motion Fusion significantly outperforms the standard  baseline while reducing the computational overhead by 50\%. Notably, our approach achieves a substantial gain in the stringent vIoU@0.5 metric. This demonstrates that by adaptively focusing on sparse  sampling points rather than dense global dependencies, our method effectively mitigates spatial-temporal noise and provides superior localization precision for STVG task.

\begin{table}[t]
  \centering
  \small 
  
  \begin{minipage}{0.48\textwidth}
    \centering
    \caption{\textbf{Ablation on SGTA Design.} Evaluated on the HC-STVGv1 dataset. TA and SG denote Temporal Attention and Semantic Guide, respectively.}
    \label{tab:ablation_sgta}
    \resizebox{\textwidth}{!}{ 
      \begin{tabular}{lcccc}
        \toprule
        Configuration & tIoU & vIoU & vIoU@0.3 & vIoU@0.5 \\
        \midrule
        \textbf{Full SGTA} & \textbf{53.7} & \textbf{40.0} & \textbf{64.0} & \textbf{39.5} \\
        w/o TA.  & 52.8 & 39.3 & 63.1 & 38.7 \\
        w/o SG.  & 53.1 & 39.5 & 63.4 & 39.0 \\
        \bottomrule
      \end{tabular}
    }
  \end{minipage}
  \hfill 
  \begin{minipage}{0.48\textwidth}
    \centering
    \caption{\textbf{Ablation on fusion mechanism.} Deformable Semantic-Motion Fusion achieves superior performance with significantly lower complexity.}
    \label{table:fusion_comparison}
    \resizebox{\textwidth}{!}{
      \begin{tabular}{lcccc}
        \toprule
        Mechanism & tIoU & vIoU & vIoU@0.5 & GFLOPs \\ 
        \midrule
        Standard SA & 53.3 & 39.8 & 36.5 & 516 \\
        \textbf{Deform. SA} & \textbf{53.7} & \textbf{40.0} &  \textbf{39.5} & \textbf{260} \\ 
        \bottomrule
      \end{tabular}
    }
  \end{minipage}

\end{table}

\subsection{Qualitative Results}
\begin{figure}[tb]
  \centering
  \includegraphics[width=\textwidth]{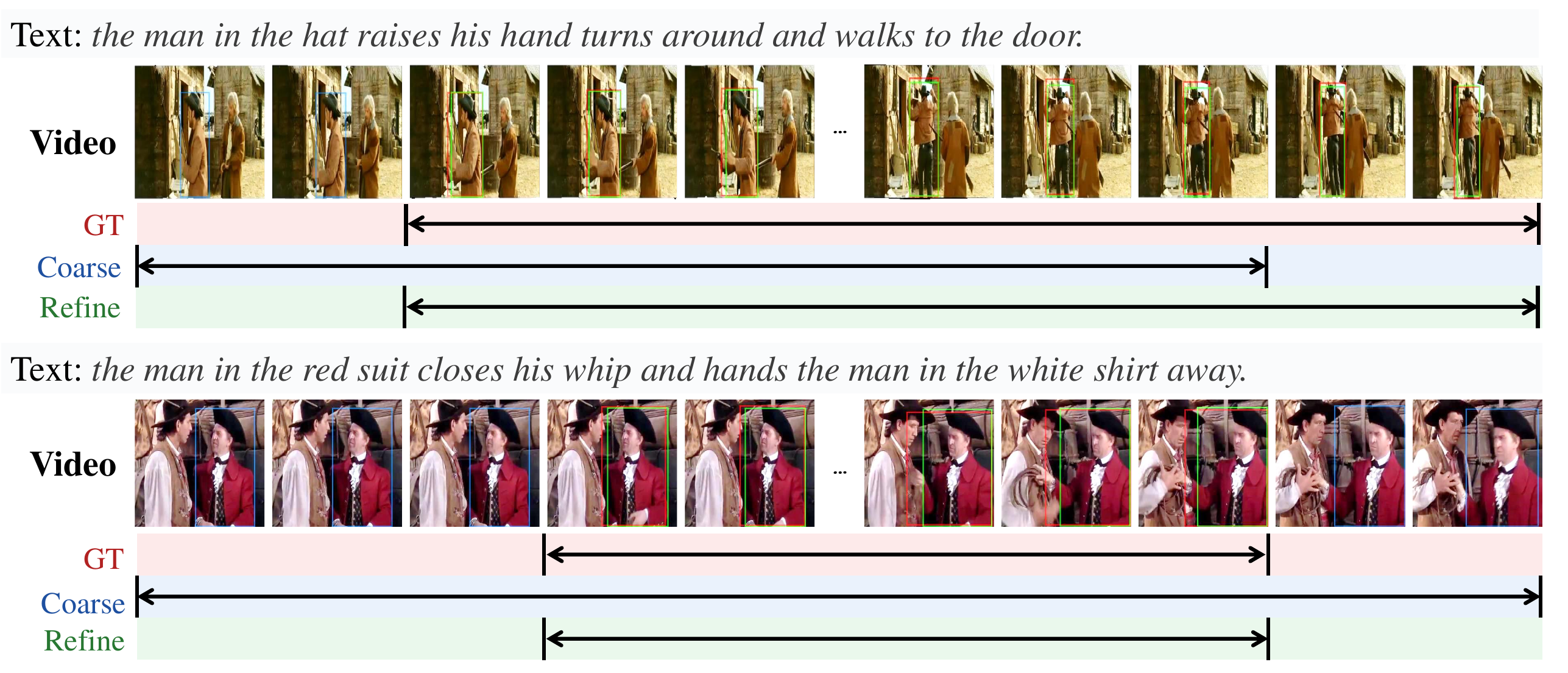}
  \caption{\textbf{Qualitative results of ScanFocus.} The coarse stage often generates ambiguous boundaries. Our refine stage successfully corrects these initial proposals to align precisely with the Ground Truth.}
  \label{fig:visualization}
\end{figure}

To intuitively demonstrate the effectiveness of our coarse-to-fine framework, we present qualitative grounding results in Fig.~\ref{fig:visualization}. As illustrated, the initial coarse stage often produces ambiguous temporal boundaries. Guided by explicit inter-frame reasoning within the local boundary focus, our refine stage successfully corrects these initial predictions by expanding or trimming the temporal windows, achieving more precise alignment with the Ground Truth. For more visualization examples, please refer to the Supplementary Material.

\section{Conclusion}
We propose \textbf{ScanFocus}, a novel coarse-to-fine framework that resolves temporal ambiguity in STVG by decoupling the task into a global spatio-temporal scan and a local boundary focus. To capture suppressed high-frequency cues and model explicit inter-frame dependencies, we introduce SGTA for fine-grained temporal modeling with dense boundary sampling. Extensive experiments across three benchmarks demonstrate that our method significantly outperforms existing SOTA approaches, establishing a highly effective  paradigm for STVG. 

\section*{Acknowledgements}
This work is supported by the National Natural Science Foundation of China under Nos.~62276061 and 62436002. This work is also supported by Research Fund for Advanced Ocean Institute of Southeast University (Major Program MP202404).

\bibliographystyle{splncs04}
\bibliography{main}
























\end{document}